\documentclass[10pt,twocolumn,letterpaper]{article}

\usepackage{cvpr}

\usepackage{mypackages}
\pdfoutput=1

\usepackage{times}
\usepackage{epsfig}
\usepackage{graphicx}
\usepackage{amsmath,amssymb} 
\usepackage{color}

\def\myalgoname{WISeR\@\xspace}

\def\vaImWidth{0.32\linewidth}
\def\vaImVisualResWidth{0.195\linewidth}
\def\samplesImWidth{0.33\linewidth}
\def\vspaceafterfigure{\vspace{-0.5em}}
\def\vspaceaftertable{\vspace{-0.6em}}

\def\xvec{\mathbf{x}}
\def\xvecl{\xvec_l}
\def\xvecll{\xvec_{l+1}}
\def\mappingfun{\mathcal{M}}
\def\residualfun{\mathcal{F}}
\def\residualpar{\mathcal{W}}
\def\residualparl{\residualpar_l}
\def\residualparlmatrixk{\mathbf{W}_{l,k}}
\def\residualproj{\mathbf{P}}
\def\mybn{\texttt{BN}}
\def\myrelu{\texttt{ReLU}}
\def\myconv{\texttt{Conv}}

\cvprfinalcopy 

\def\cvprPaperID{2641} 

\ifcvprfinal\fi
\begin{document}

\title{Wide-Slice Residual Networks for Food Recognition}

\author{Niki Martinel\\
	University of Udine\\
	{\tt\small niki.martinel@uniud.it}
	\and
	Gian Luca Foresti\\
	University of Udine\\
	{\tt\small gianluca.foresti@uniud.it}
	\and
	Christian Micheloni\\
	University of Udine\\
	{\tt\small christian.micheloni@uniud.it}
}

\maketitle

\begin{abstract}
Food diary applications represent a tantalizing market.
Such applications, based on image food recognition, opened to new challenges for computer vision and pattern recognition algorithms.
Recent works in the field are focusing either on hand-crafted representations or on learning these by exploiting deep neural networks.
Despite the success of such a last family of works, these generally exploit off-the shelf deep architectures to classify food dishes.
Thus, the architectures are not cast to the specific problem.
We believe that better results can be obtained if the deep architecture is defined with respect to an analysis of the food composition.
Following such an intuition, this work introduces a new deep scheme that is designed to handle the food structure.
Specifically, inspired by the recent success of residual deep network, we exploit such a learning scheme and introduce a slice convolution block to capture the vertical food layers.
Outputs of the deep residual blocks are combined with the sliced convolution to produce the classification score for specific food categories.
To evaluate our proposed architecture we have conducted experimental results on three benchmark datasets.
Results demonstrate that our solution shows better performance with respect to existing approaches (\eg, a \textit{top--1} accuracy of $90.27\%$ on the Food-101 challenging dataset).
\end{abstract}


\section{Introduction}
The recent advent of deep learning technologies has achieved successes in many visual perception tasks such as object and action recognition, image segmentation, visual question answering \etc~\cite{Simonyan2015,Szegedy2015,He2016,Feichtenhofer2016,Deng2016,Zhang2015a,Zhu2016,Hendricks2016,Hu2015}.
Yet the status quo of computer vision and pattern recognition is still far from matching human capabilities, especially when it comes to classifying an image whose intra-category appearance might present more differences than its inter-category counterparts. 
This is the case of food recognition where a particular food dish may be prepared in thousands of different ways, yet it is essentially the same food.
Reaching the final objective of food diary applications by solving the food recognition and calories estimation problems would be highly beneficial to tackle the rapid increase of diseases related to excessive or wrong food intake~\cite{who2015fs}.


Even if we relax the objective and focus only on the food recognition task, we still have to address a tough problem with many specific challenges.
Intra-class variation is a strong source of uncertainty, since the recipe for the same food can vary depending on the location, the available ingredients and, last but not least, the personal taste of the cook.
On the other hand, different foods may look very similar (\eg, soups where the main ingredients are hidden below the liquid level), thus inter-class confusion is a source of potential problems too.
Despite such issues, a quick, partial view, of a food image is often sufficient for a human to recognize the food dish.
This remarkable ability inevitably tells us that food images have distinctive properties that made task tractable, regardless the non-trivial challenges.

\begin{figure}[t]
	\centering
	\includegraphics[width=1\linewidth]{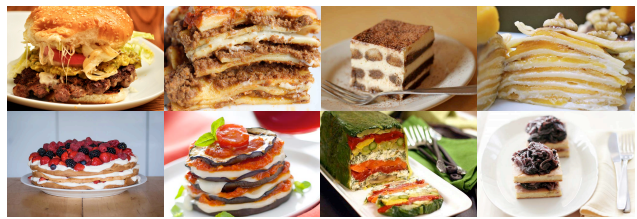}
	\caption{The food recognition problem is characterized by large intra class variations.
	However, some dishes present a spatial structure which has been not considered so far.}
	\label{fig:problem}
\end{figure}

Methods tackling the food recognition task largely explored hand-crafted image representations based on~\textit{a priori} knowledge of the problem (\eg,~\cite{Matsuda2012,Farinella2014a,Qi2014}).
Such a knowledge yields to encouraging results (\eg,~\cite{Farinella2014a,Bossard2014,Martinel2016} obtained considering combinations of different features (\eg, color, shape, spatial relationships, \etc).
Recently, by riding the deep learning wave of enthusiasm, works investigated the possibility of learning specific food image representation~\cite{Kawano2014b,Liu2016,Hassannejad2016,Chen2016,Myers2016}.
Such approaches generally obtained better performance thanks to a mere application of off-the-shelf architectures to the problem.
Thus, existing approaches neglected the design of a proper architecture which considers the specific problem challenges.
\emph{This motivates the development of a novel architecture that is defined following an analysis of the food composition.}

The proposed WIde-Slice Residual Network (\myalgoname) solution builds upon such an idea by proposing a deep learning architecture which aims to capture the food structure.
Our key intuition is that, regardless the exploited ingredients and the final presentation, many dishes are largely characterized by vertical food layers (see~\figurename~\ref{fig:problem} for same examples).
Thus we propose to leverage such a vertical structure to introduce the \textit{slice convolution layer}.
However, since not all the food dishes present such a structure, we also exploit a large residual learning architecture to obtain a generic food representation.
This, together with representation obtained from the slice convolution, is enclosed in single architecture to emit the food classification. 

\textbf{Contributions:}
Concretely, our contributions are:
\begin{enumerate*}[label=(\roman{*})]
\item We propose a novel convolutional layer that captures the vertical structure of food dishes;
\item By combining the features detected through such layer with a stack of residual learning blocks we obtain a good representation for food dishes which do not show a specific structure;
\item Significantly enlarge the number of feature maps per convolution layer to tackle the diminishing feature reuse issue in deep residual networks~\cite{Srivastava2015}, thus to improve the representational power of the learned feature detectors.
\end{enumerate*}

Results on three benchmark datasets show that, by combining such three ingredients together, our approach performs better than existing works in the field.

\section{Related Work}
The food recognition problem is a recent field of research in computer vision and pattern recognition.
In the following, we review the most relevant work to our approach.

\noindent\textbf{Food Recognition.}
During the last few years, the topic of food recognition for health-oriented applications has gained increasing popularity.
One of the earliest work in the field appeared in~\cite{Yang2010}, where authors proposed to study the spatial relationships between different food ingredients.
A Semantic Texton Forest was exploited to segment each image into eight different types of ingredient, then pairwise statistics were exploited to compute a multi-dimensional histogram, later classified with an SVM.
Starting from such a work, plenty of research has been carried out to find the optimal hand-crafted representation for food recognition.
In~\cite{Farinella2014}, the Maximum Response Filter Bank (MR) was used in a Bag of Textons (BoT) scheme. 
Such a representation, combined with color descriptors in a nearest neighbor approach~\cite{Farinella2014} demonstrated that both such clues are relevant for the task.
The idea of exploiting multiple features was recently brought to the limit by considering as many features as possible and limiting their importance through an ensemble fusion scheme~\cite{Martinel2016}.

\noindent\textbf{On Mobile Devices.}
Other works were specifically tuned to work on smartphones or other devices with limited capabilities.
DietCam~\cite{Kong2012} assesses daily food intakes by fusing the classification of a SIFT-based Bag of Visual Words (BoW) and a nearest-neighbour-based best match search.
Similarly, in~\cite{Kawano2013}, a BoW-based scheme was applied to obtain an encoding of visual features extracted from segmented food items.
A segmentation scheme was also explored by the FoodCam app~\cite{Kawano2014}.
It performs HoG and color patch feature encoding via Fisher Vector (FV)~\cite{Sanchez2013} and classify those with a one-vs-rest linear SVM.
Context information regarding the location from where the food picture was taken, together with additional information about the restaurant, had been exploited in~\cite{Bettadapura2015a}.
Such information, coupled with a multiple kernel learning scheme applied on different visual features yield to the food image classification.

\noindent\textbf{Calories Estimation.}
The idea of exploiting restaurant information was also explored in~\cite{Myers2016}.
This, fused with the result of food detectors and volume estimation was used to directly tackle the calories estimation problem.
The same issue was also explored in~\cite{Zhu2010}, which required manually-driven image segmentation and camera calibration.
Finally, by viewing the problem from a different perspective, in~\cite{Matsuda2012}, authors proposed to recognizing multiple food items appearing in the same picture.
Outputs of different region detectors were fused to identify different foods, which were later classified using texture features and an SVM.
Pairwise co-occurrence statistics~\cite{Qi2014} were also exploited to improve performance. 

While being able to reach good recognition performance (\eg,~\cite{Qi2014,Bettadapura2015a,Martinel2016}) such works rely on the available a priori knowledge of the problem.
Thus, they do not consider that the exploited hand-crafted feature representation may not be the optimal one for the classification objective.
Differently from all such schemes our approach does not hinge on the manual selection of such features.
It addresses the problem by learning and exploiting only the optimal features for classification.

\noindent\textbf{Deep Learning for Food Recognition.}
A small number of studies have explored the applicability of deep neural networks to food recognition.
Specifically, in~\cite{Kawano2014b} the output of an AlexNet-style architecture pretrained on ImageNet plus additional 1000 disjoint food-related categories was combined with traditional hand-crafted features embedded through FV.
Deeper architectures following the Inception~\cite{Szegedy2015} structure were exploited in~\cite{Liu2016,Hassannejad2016}.
In details, in~\cite{Liu2016}, the Inception module is modified by introducing $1\times 1$ convolutional layers to reduce the input dimension to the next layers. 
Similarity, in~\cite{Hassannejad2016}, the convolutional layers of the Inception v3 network were specifically modified to improve the computational efficiency.
Finally, a modified VGG-16 net~\cite{Simonyan2015} together with a multi-task loss was exploited in~\cite{Chen2016} to tackle the problems of food and ingredient recognition.
A Conditional Random Field was then employed to tune the probability distribution of ingredients.
Due to the scarcity of food images, all such works exploited networks which were pretrained on ImageNet, then finetuned to classify food categories.

Our work has several key differences with the aforementioned works:
First and foremost, we have designed a specific convolutional layer that handles the structural peculiarities of some food dishes.
Second, the methods above~\cite{Kawano2014b,Liu2016,Hassannejad2016,Chen2016} consider classic off-the-shelf deep learning architectures.
Our work is the first work exploiting residual learning~\cite{He2016} for food recognition. 
Moreover, our model uses a significantly large number of feature maps per convolution layer to tackle the diminishing feature reuse issue in deep residual networks~\cite{Srivastava2015}.

\section{Wide-Slice Residual Networks}
Our goal is to take a single-shot of a food dish and output the corresponding food category.
The proposed model aims to achieve such an objective by discovering the structural peculiarities of the image by combining a slice convolution layer with residual learning.

\subsection{Architecture}
\begin{figure}[t]
	\centering
	\includegraphics[width=.74\linewidth]{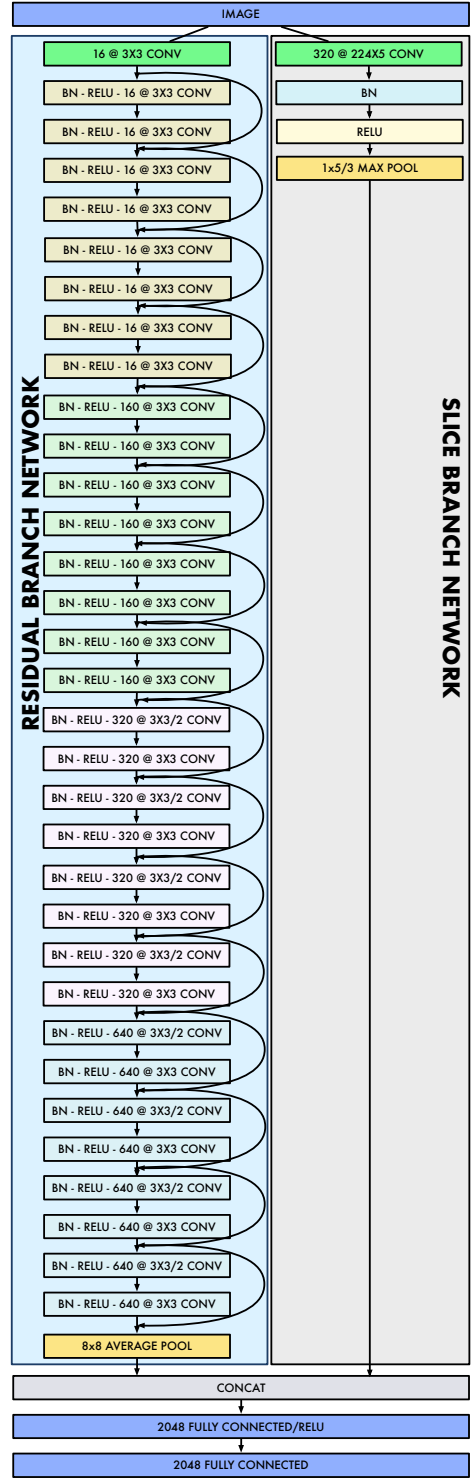}
	\caption{Proposed~\myalgoname architecture consisting of two branches: a residual network branch (Sec.\ref{sec:residual_branch}), and a slice branch network with slice convolutional layers (Sec.\ref{sec:slice_branch}).
	The output of the two branches in fused via concatenation, then fed to the two fully connected layers to emit the food classification prediction.}
	\label{fig:wisernet}
	\vspace{-3.5em}
\end{figure}
As shown in \figurename~\ref{fig:wisernet}, the model consists of a single deep network with two main branches: a residual network branch (Sec.~\ref{sec:residual_branch}), and a slice network branch with a slice convolutional layer (Sec.~\ref{sec:slice_branch}).
The residual network encodes generic visual representations of food images. 
The slice network specifically captures the vertical food layers.
Features extracted from the two branches are then concatenated and fed to the fully connected layers that emit a classification prediction. 
We now describe each of these two branches in more details.

\subsection{Residual Network Branch}
\label{sec:residual_branch}

\subsubsection{Residual Learning}
Since the breakthrough paper on extremely deep neural networks first appeared in~\cite{He2015} and later published in~\cite{He2016} --which won the ILSVRC and MSCOCO 2015 competitions, a surge of effort has been dedicated on exploring residual learning in such architectures. 
The idea behind residual learning is very simple yet has been shown to be extremely effective~\cite{He2016} in solving optimization issues that affects the process of learning the parameters of very deep neural networks (\eg, with more than 20 layers).

Everything starts from the assumption that given an input $\xvec$, a shallow network with few stacked non-linear layers can approximate a mapping function $\mappingfun(\xvec)$.
On the basis of such an assumption, it is reasonable to hypothesize that a network with the same structure can approximate the residual function $\residualfun(\xvec) = \mappingfun(\xvec) - \xvec$ (given that the input and output have the same dimensionality).
While, either learning an approximation of the mapping function $\mappingfun(\xvec)$ or the residual function $\residualfun(\xvec)$ is feasible, the ease of such a process is significantly different.
Indeed, as demonstrated in~\cite{He2016}, deep networks trained to approximate the mapping function $\mappingfun(\xvec)$ suffer from a degradation that does not appear in networks trained on approximating the residual function $\residualfun(\xvec)$.
This opens to the success of residual learning for very deep networks.

\subsubsection{Wide Residual Blocks}
\begin{figure}[t]
	\centering
	\begin{subfigure}[]{.41\linewidth}
		\includegraphics[width=1\linewidth]{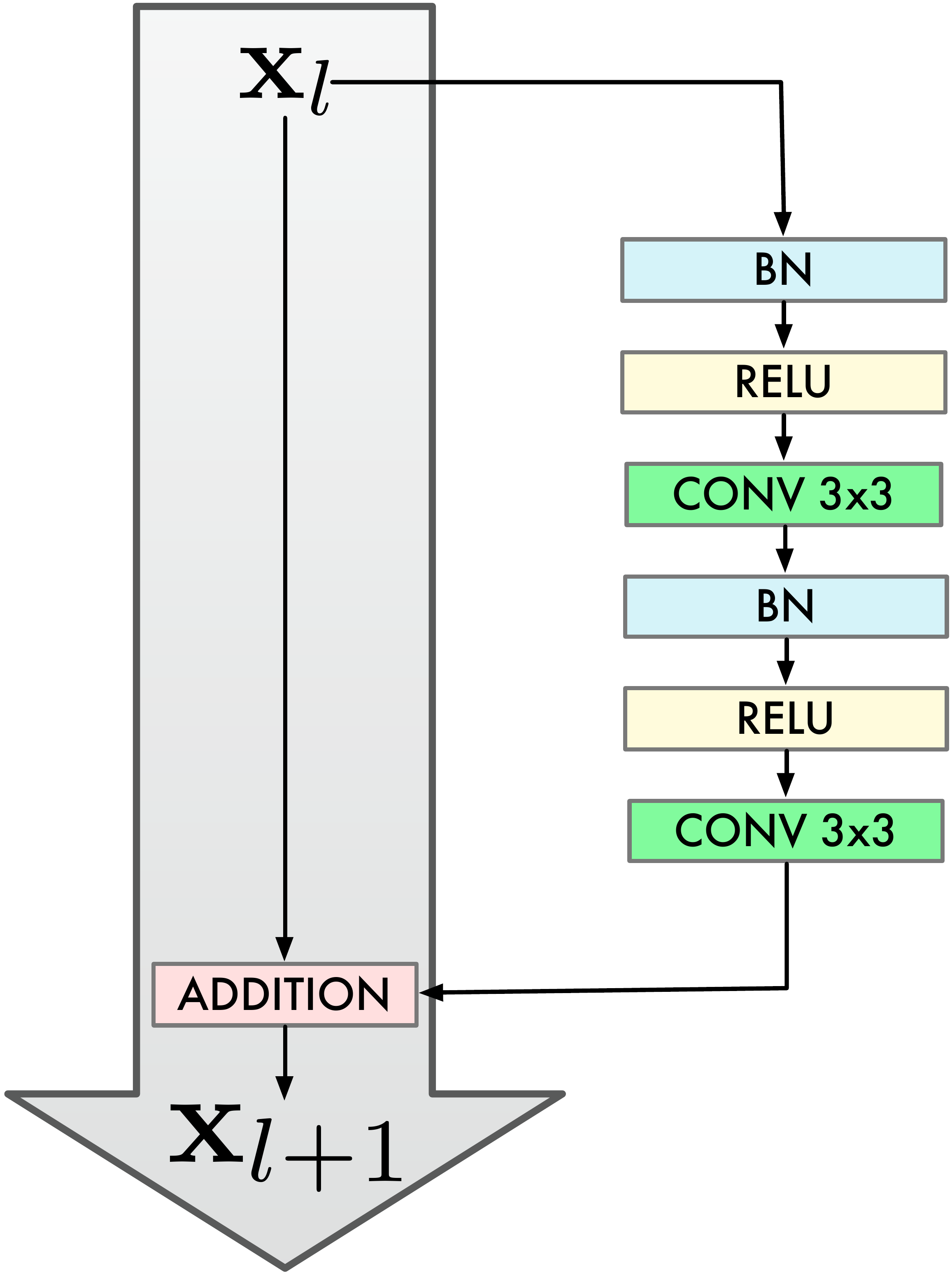}
		\caption{}
		\label{fig:resblock}
	\end{subfigure}
	\begin{subfigure}[]{.48\linewidth}
		\includegraphics[width=1\linewidth]{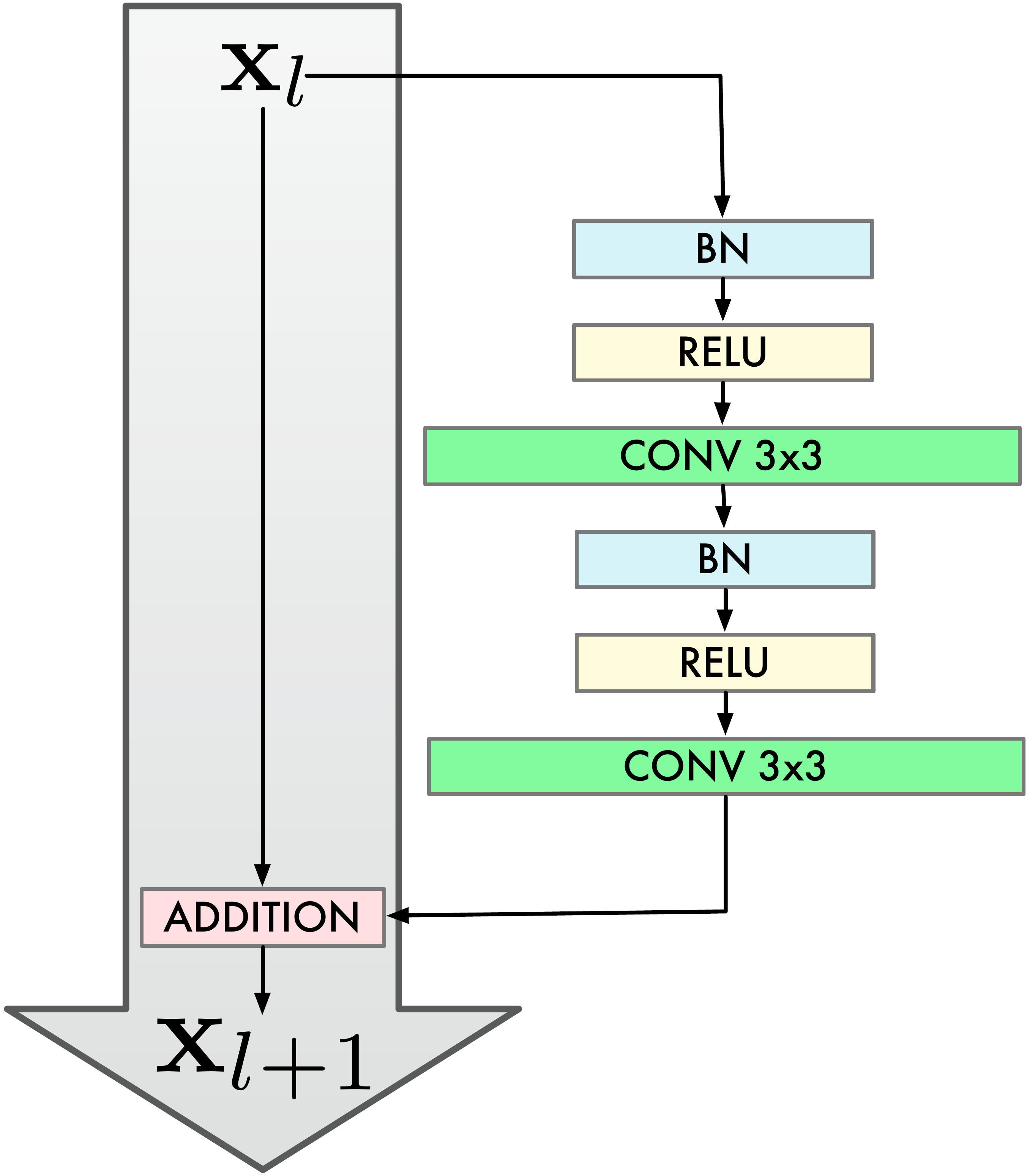}
		\caption{}
		\label{fig:resblock_wide}
	\end{subfigure}
	\caption{Graphical representation of~\subref{fig:resblock} Basic Residual Blocks and~\subref{fig:resblock_wide} Wide Residual Blocks.
	By expanding the number of convolution kernels (\ie, widening), the number of parameters to learn increases, hence the networks has more capacity.}
	\label{fig:resblocks}
\end{figure}
Following the methodology in~\cite{He2016}, we exploit residual learning every few stacked layers.
Formally, we let a residual block with identity mapping~\cite{He2016a} be represented as
\begin{equation}
\label{eq:residual_block}
\xvecll = \xvecl + \residualfun(\xvecl, \residualparl)
\end{equation}
where $\xvecl$, $\xvecll$ and $\residualparl = \{\residualparlmatrixk | k=1,\ldots,K\}$ represent the input, the output and the set of parameters associated with the $l$-th residual block, respectively.
$K$ denotes the number of layers in a residual block ($K=2$, in our case).
The residual learning objective is to find the parameters $\residualparl$ that best approximate the function $\residualfun(\xvecl, \residualparl)$.

Before going further it is important to emphasize a few relevant ingredients regarding eq.(\ref{eq:residual_block}):
\begin{enumerate}[label=\roman{*})]
\item neither extra parameter nor computation complexity is introduced (except for the negligible addition performed on feature maps, channel by channel);
\item the function $\residualfun(\xvecl, \residualparl)$ is very flexible and can represent both fully connected and convolutional layers;
\item if the dimensionalities of $\xvecl$ and $\residualfun(\xvecl, \residualparl)$ are different (\eg, when varying the number of feature maps), a linear projection $\residualproj$ can be exploited by the shortcut connections to match the dimensions (\ie, eq.(\ref{eq:residual_block}) becomes $\xvecll = \residualproj\xvecl + \residualfun(\xvecl, \residualparl)$).
\end{enumerate}

Armed with eq.(\ref{eq:residual_block}), as shown in~\figurename~\ref{fig:resblocks}, we followed the recommendations in~\cite{He2016a} and adopted the batch normalization (\mybn) and ReLU (\myrelu) layers as ``pre-activations'' for the convolutional layers (\myconv).
Then, to increase the representational power of a residual block we shared the same idea as~\cite{Zagoruyko2016} and widen the convolutional layers by significantly increasing the number of feature maps.
This has been shown to be able to tackle the diminishing feature reuse problem~\cite{Srivastava2015} and improve performance of residual networks compared to increasing their depth.

\subsection{Slice Network Branch}
\label{sec:slice_branch}

\subsubsection{Slice Convolution}
\begin{figure}[t]
	\centering
	\begin{subfigure}[]{.48\linewidth}
		\includegraphics[width=1\linewidth]{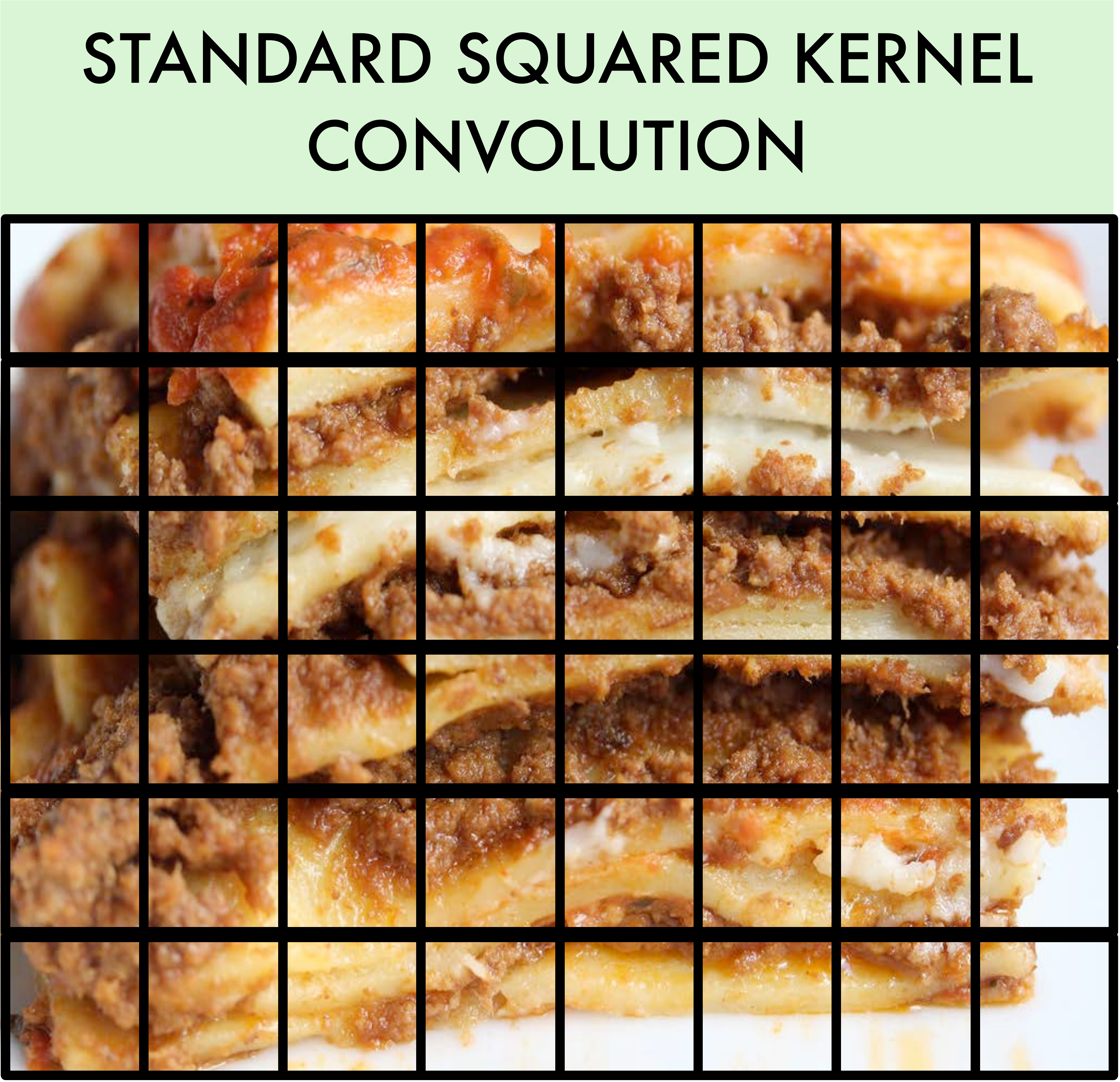}
		\caption{}
		\label{fig:conv}
	\end{subfigure}
	\begin{subfigure}[]{.48\linewidth}
		\includegraphics[width=1\linewidth]{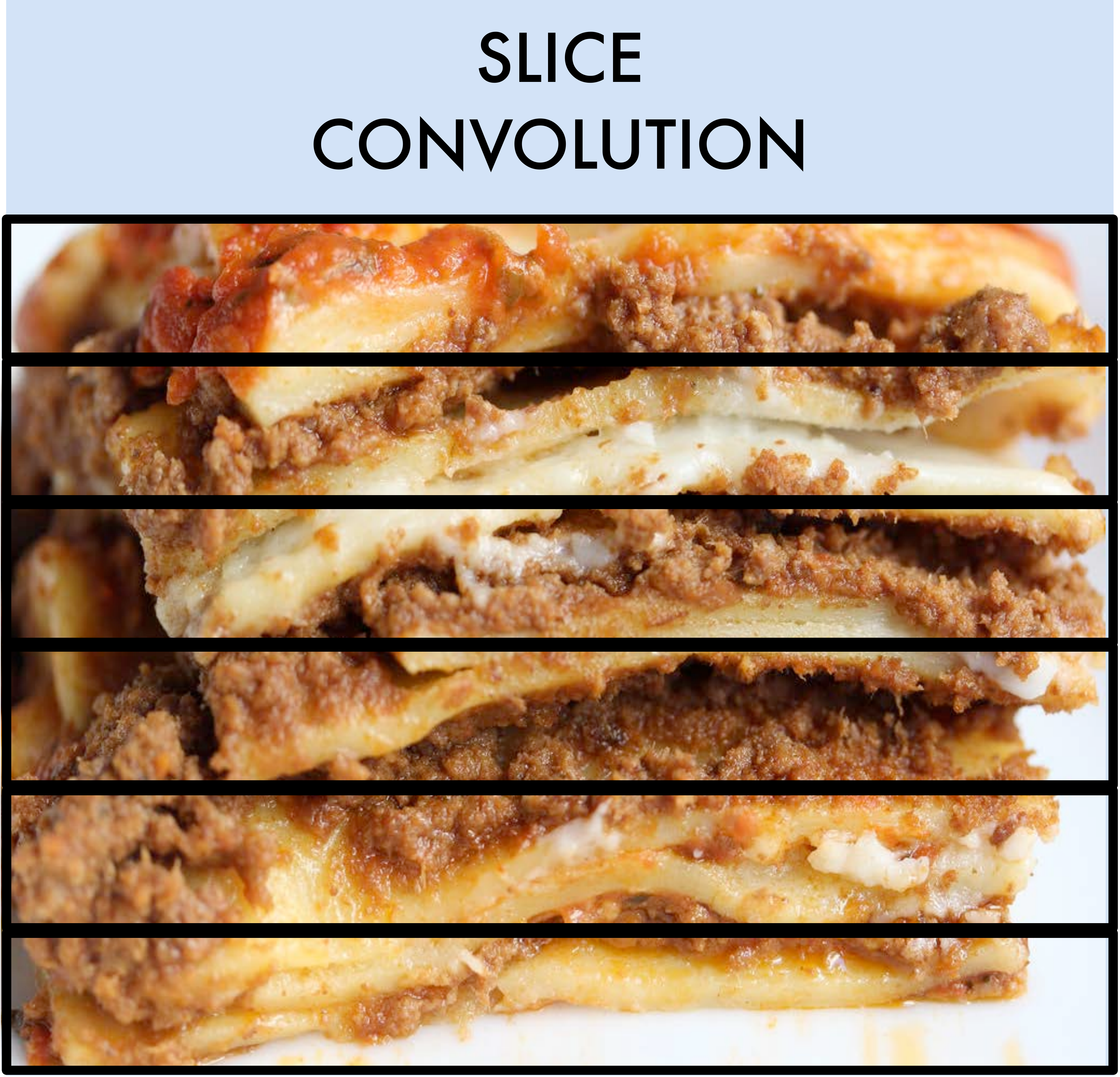}
		\caption{}
		\label{fig:slicedconv}
	\end{subfigure}
	\caption{\subref{fig:conv} Standard squared convolutional kernel commonly used in deep learning architectures for food recognition.~\subref{fig:slicedconv} Proposed slice convolutional kernel aiming to capture the vertical layer structure of some food dishes.}
	\label{fig:conv_kernels}
\end{figure}
Common deep learning architectures (\eg,~\cite{Simonyan2015,Szegedy2015,He2016,He2016a}) exploit squared kernels to detect and extract relevant image features (see~\figurename~\ref{fig:conv}).
Same occurs when such architectures are applied to the food recognition task (\eg,~\cite{Kawano2014b,Liu2016,Hassannejad2016,Chen2016}).
By doing this, existing approaches do not directly consider the vertical traits of some food dishes.
We believe that, while such a vertical structure can be captured via the deep hierarchy, a specific food layer detector can be extremely useful when it comes to classify food dishes that present such a peculiarity.

For such a purpose, we propose to exploit a slice convolution (see~\figurename~\ref{fig:slicedconv}).
It will learn the parameters of a convolution kernel that has the same width as the input image.
In such a way, it will act as a vertical layer feature detector.

\begin{figure*}[t]
	\centering
	\begin{subfigure}[]{\samplesImWidth}
		\includegraphics[width=\linewidth]{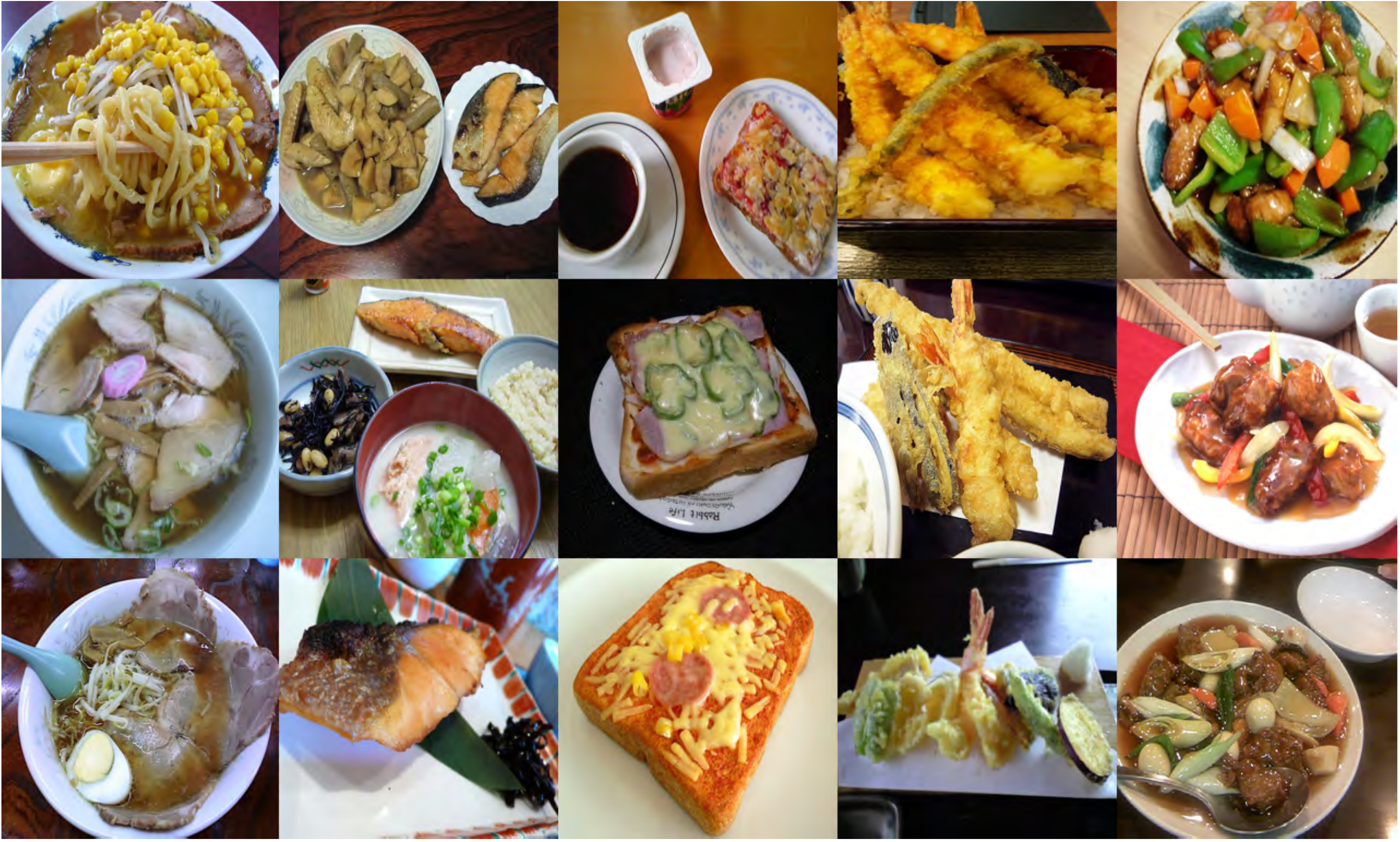}
		\caption{}
		\label{fig:samples_food100}
	\end{subfigure}
	\begin{subfigure}[]{\samplesImWidth}
		\includegraphics[width=\linewidth]{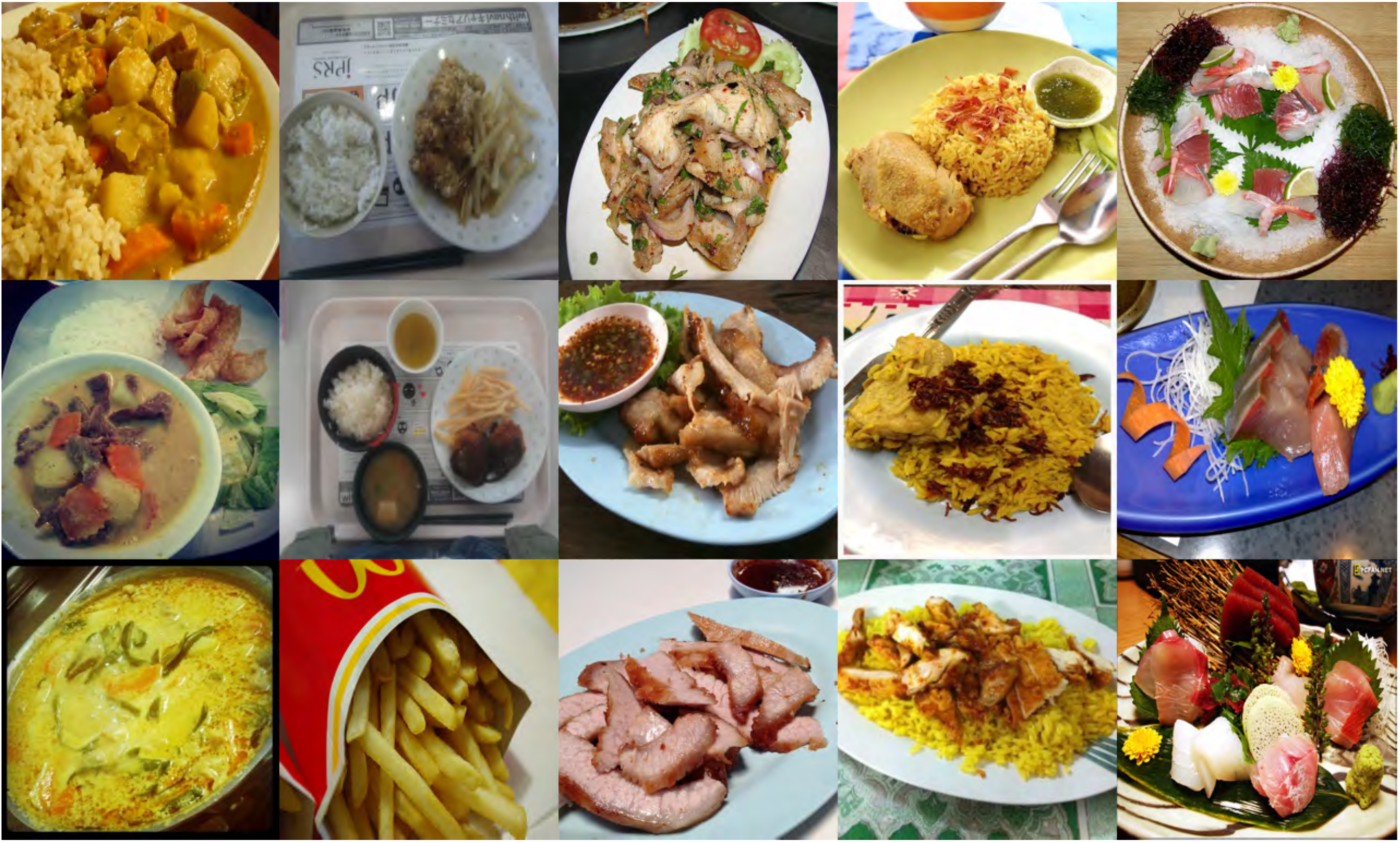}
		\caption{}
		\label{fig:samples_food256}
	\end{subfigure}
	\begin{subfigure}[]{\samplesImWidth}
		\includegraphics[width=\linewidth]{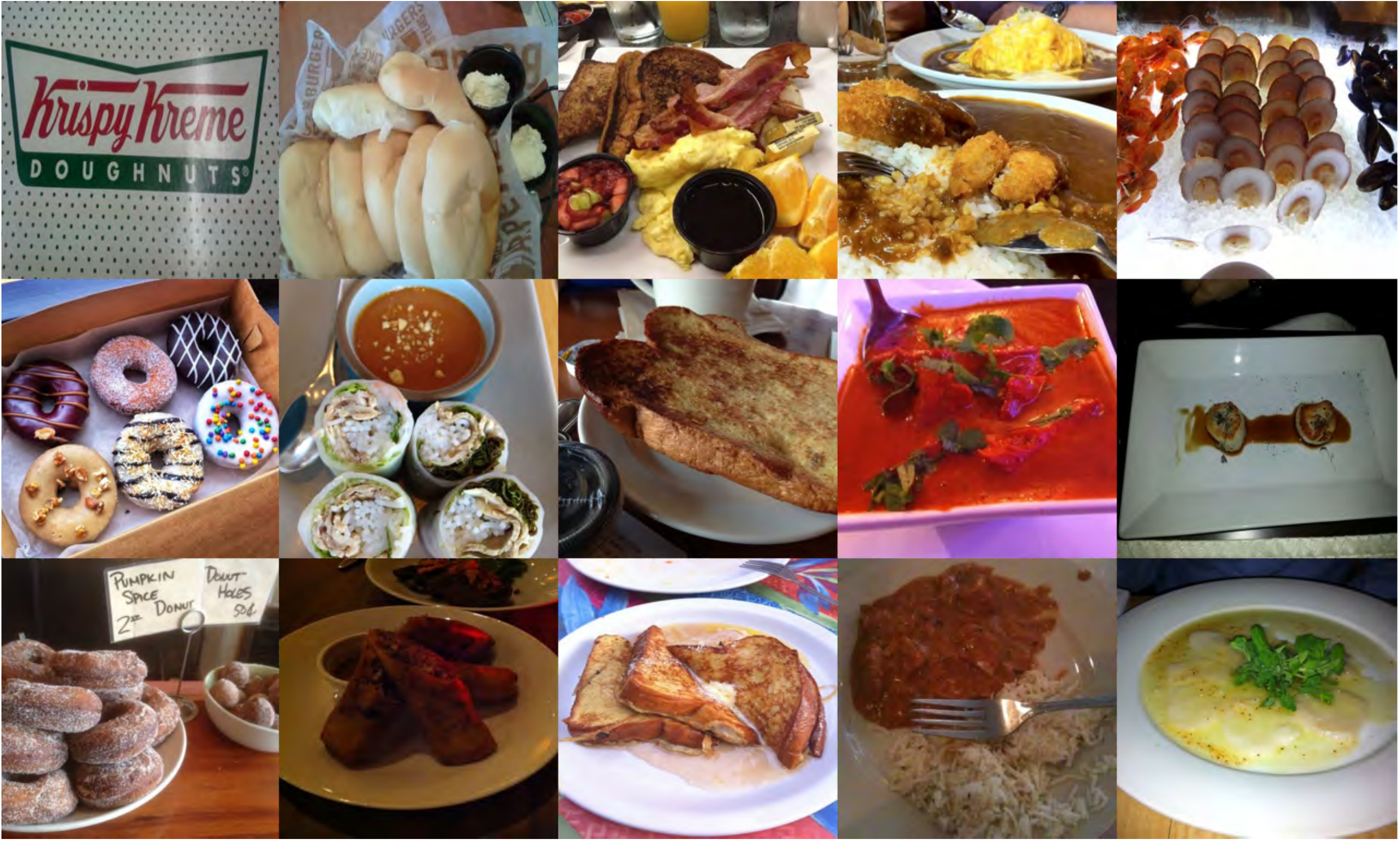}
		\caption{}
		\label{fig:samples_food101}
	\end{subfigure}
	\caption{Five different food dishes appearing in the~\subref{fig:samples_food100} UECFood100,~\subref{fig:samples_food256} UECFood256, and~\subref{fig:samples_food101} Food-101 datasets.
		The three rows highlight the strong intra-class variations for the same food dish. 
		\textit{(Best viewed in color)}}
	\label{fig:dataset_samples}
	\vspaceafterfigure
\end{figure*}

\subsubsection{Slice Pooling}
For a specific vertically structured food category, it is not guaranteed that the vertical layers appears in the same position.
Thus, the output of the slice convolution might be different depending on the location of such vertical traits.
To tackle this issue we perform max pooling on vertically elongated windows.
As a result we expect that a specific food layer is detected within a certain vertical location.

\section{Experimental Results}
First, we describe the selected datasets and the evaluation protocol.
This is followed by a discussion of experimental and design selections. 
Then, we present the comparisons with existing methods to demonstrate the superior performance of \myalgoname, followed by concluding remarks.

\subsection{Datasets}
To validate the proposed \myalgoname approach, results on three benchmark datasets for food recognition have been computed.
These have been selected on the basis of the different challenges they carry.

\noindent \textbf{UECFood100\footnote{\url{http://foodcam.mobi/dataset100}}.}
The UECFood100 dataset~\cite{Matsuda2012} contains 100 different food categories for a total of approximately 14'000 images.
Images acquired by mobile cameras contain the most popular Japanese food dishes.
Since the dataset has been conceived to address a real-world challenge, the acquired pictures may contain more than a single food dish.
Therefore, this dataset is useful to understand if the approach is able perform food localization before classification, hence if it focuses on the relevant image details.


\noindent \textbf{UECFoo256\footnote{\url{http://foodcam.mobi/dataset256}}.}
The UEC-Food256~\cite{Kawano2014c,Kawano2014d} is a newly-constructed food image dataset, which has been build on the idea that number of food categories in existing datasets is not enough for practical use.
Authors exploited knowledge on food of other countries and leveraged on existing categories to extend the UECFood100 dataset.
The so obtained dataset contains 256 different foods dishes which are represented in about 32'000 images.
As for the UECFood100 dataset, multiple food dishes can appear in a same image.
With this dataset we aim to evaluate our approach on classifying a large number of challenging classes. 

\noindent \textbf{Food-101\footnote{\url{http://www.vision.ee.ethz.ch/datasets/food-101/}}.}
The food-101 dataset~\cite{Bossard2014} consists of real-pictures of the 101 most popular dishes that appeared on \url{foodspotting.com}.
On purpose, the images have not been selected and checked by human operator, hence the training set contains 75750 images with intense colors and sometimes wrong labels.
Additionally, 250 test images have been collected for each class, and have been manually cleaned.
The dataset has a total of 101'000 realworld images, including very diverse but also visually and semantically similar food classes.
This allows us to validate our approach on a large dataset build with weakly labeled data.

\subsection{Evaluation Protocol}
Evaluation of food recognition approaches (\eg,~\cite{Matsuda2012,Kawano2014,Farinella2014a,Bettadapura2015a}) is generally performed by showing the~\textit{Top-−1} recognition accuracy.  
In addition to that, we also report on the~\textit{top-−5} criterion as generally considered when providing the results achieved by deep neural networks.

For the UECFood256 and Food-101 datasets, we used the provided splits.
Since the UECFood100 dataset does not come with such a feature, we evaluated the performance of our approach using the same protocol in~\cite{Kawano2014b,Hassannejad2016}, hence randomly partitioned the dataset into two subsets using $80\%$ of the images for training and the rest for testing.

Notice that, the performance achieved by the existing methods have been taken from the corresponding works or have been directly provided by the authors.

\begin{table}[t]
\small
\centering
\caption{\textit{Top--1} and \textit{Top--5} performance on the UECFood100 dataset.
	First 4 rows show the results achieved by using methods adopting hand-crafted features.
	Next 11 rows show the performance obtained by deep learning-based approaches on the ground-truth cropped images.
	Last 2 rows depict the results obtained considering images having more than a single food class (\ie, no ground truth is exploited).
Best results is highlighted in boldface font.}
\label{tab:uecfood100_comparison}
\begin{tabulary}{1\linewidth}{L{7.9em} C{2.6em}  C{2.6em} | L{8.5em}}
\toprule
\textbf{Method} & \textbf{Top-1} & \textbf{Top-5} & \textbf{Publication} \\ \midrule
MKL 				& 51.6 & 76.8 & COST2016~\cite{Liu2016} \\
FC7 				& 58.03 & 83.71 & ACMMM2016~\cite{Chen2016}\\
Extended HOG Patch-FV+Color Patch-FV(flip) & 59.6 & 82.9 & COST2016~\cite{Liu2016} \\
SELC				& 84.3 & 95.2 & CVIU2016~\cite{Martinel2016} \\
\hline \hline
DeepFoodCam		 	& 72.26 & 92.00 & UBICOMP2014~\cite{Kawano2014b}\\
AlexNet 			& 75.62 & 92.43 & ACMMM2016~\cite{Chen2016}\\
DeepFood  			& 76.3 	& 94.6  & COST2016~\cite{Liu2016}\\
FV+DeepFoodCam 		& 77.35 & 94.85 & UBICOMP2014~\cite{Kawano2014b} \\
DCNN-FOOD 			& 78.77 & 95.15 & ICME2015~\cite{Yanai2015a}\\
VGG					& 81.31 & 96.72 & ACMMM2016~\cite{Chen2016} \\
Inception V3 		& 81.45 & 97.27 & ECCVW2016~\cite{Hassannejad2016}\\
Arch-D				& 82.12 & 97.29 & ACMMM2016~\cite{Chen2016} \\
ResNet-200			& 86.25 & 98.91 & CVPR2016~\cite{He2016}\\
WRN					& 86.71 & 98.92 & BMVC2016~\cite{Zagoruyko2016}\\
\myalgoname			& \textbf{89.58} & \textbf{99.23} & Proposed \\ 
\hline \hline
DeepFood  			& 57.0 	& 83.4  & COST2016~\cite{Liu2016}\\
\myalgoname 		& 79.46 & 97.46 & Proposed \\ \bottomrule
\end{tabulary}
\vspaceaftertable
\end{table}

\subsection{Experimental and Implementation Settings}
Existing food recognition deep net-based approaches tweaks the network hyperparameters to the specific dataset (\eg,~\cite{Kawano2014b,Hassannejad2016}).
In our evaluation, we have decided not to specifically adjust them to provide a generic framework.

Following the common recipe adopted by existing approaches~\cite{Kawano2014b,Chen2016,Bettadapura2015a}, we did not train our architecture from scratch since it required more food images than all the ones that are currently available in any dataset.
We started from a WRN architecture~\cite{Zagoruyko2016} pre-trained on the ImageNet 2012 (ILSVRC2012) classification dataset~\cite{Russakovsky2015}.
Then, we added the slice convolution branch and fine-tuned the whole architecture on the selected food recognition datasets.

\noindent\textbf{Data.}
During the fine-tuning process we augment the number of dataset samples by taking $224\times 224$ random crops from images resized such that the smaller dimension is of 256 pixels.
We also exploited horizontal flipping with the scale and aspect ratio augmentation technique proposed in~\cite{Szegedy2015}.
In addition, we applied photometric distortions~\cite{Howard2013} and the AlexNet-style color augmentation~\cite{He2016}.
In testing, we considered the standard 10-crop testing~\cite{Krizhevsky2012a}.

\noindent\textbf{Optimization.}
Model training was performed via stochastic gradient descent with mini-batches containing $24$ samples.
The initial learning rate has been set to $0.01$, then updated to $0.002$ and $0.0004$, after 50k and 90k iterations respectively.
Momentum has been set to $0.9$ and a weight decay penalty of $0.0005$ had been applied to all layers.
Training has been stopped after $100$k iterations.
All the experiments have been ran using the Torch neural network framework on a multi-GPU server.

\subsection{Performance Analysis}
\label{sec:performance_analysis}

\subsubsection{State-of-the-art Comparisons}
In the following, the performances of our approach are compared to the state-of-the-art ones on the three considered benchmark datasets.

\noindent\textbf{UECFood 100.} 
Table~\ref{tab:uecfood100_comparison} shows the results achieved by existing methods and compares our approach with the top performer~\cite{Zagoruyko2016} on the UECFood100 leaderboard.
Considering ground-truth cropped images, our architecture improves the \textit{Top--1} performance of existing works specifically designed for food recognition (\ie,~\cite{Martinel2016,Chen2016}) by more than $7\%$.
Such a gap reduces to about $2.5$ percentage points if comparison is given with respect to~\cite{Zagoruyko2016}.
The~\myalgoname architecture is the only one that surpasses the $99\%$ recognition accuracy at \textit{Top--5}.

Notably, our solution shows a significant improvement over~\cite{Liu2016} (\ie, about $20\%$) when the considered images are not cropped to contain the ground truth only, but exhibit more food dishes appearing at the same time.

\noindent\textbf{UECFood 256.}
\begin{table}[t]
\small
\centering
\caption{\textit{Top--1} and \textit{Top--5} performance on the UECFood 256 dataset.
	First 3 rows show the results obtained by using methods adopting hand-crafted features.
	Next 7 rows show the performance obtained by deep learning-based approaches on the ground-truth cropped dataset images.
	Last 2 rows depict the results obtained considering input images having more than a single food class (\ie, no ground truth is exploited).
	Best result is highlighted in boldface font.}
\label{tab:uecfood256_comparison}
\begin{tabulary}{1\linewidth}{L{8.3em} C{2.5em} C{2.5em} | L{8.5em}}
\toprule
\textbf{Method} & \textbf{Top-1} & \textbf{Top-5} & \textbf{Publication} \\ \midrule
RootHOG-FV		& 36.46 & 58.83 & UBICOMP2014~\cite{Kawano2014b} \\
Color-FV		& 41.60 & 64.00 & UBICOMP2014~\cite{Kawano2014b} \\
Color-FV+HOG-FV	& 52.85 & 75.51 & UBICOMP2014~\cite{Kawano2014b} \\
\hline \hline
DeepFoodCam		& 63.77 & 85.82 & UBICOMP2014~\cite{Kawano2014b} \\
DeepFood 		& 63.8 	& 87.2	& COST2016~\cite{Liu2016}\\
DCNN-FOOD 		& 67.57 & 88.97 & ICME2015~\cite{Yanai2015a}\\
Inception V3 	& 76.17 & 92.58 & ECCVW2016~\cite{Hassannejad2016}\\
ResNet-200		& 79.12	& 93.00	& CVPR2016~\cite{He2016}\\
WRN				& 79.76 & 93.90	& BMVC2016~\cite{Zagoruyko2016}\\
\myalgoname		& \textbf{83.15}	& \textbf{95.45}	& Proposed \\ 
\hline \hline
DeepFood 		& 54.7 	& 81.5	& COST2016~\cite{Liu2016}\\
\myalgoname 	& 72.71 & 93.78	& Proposed \\
\bottomrule
\end{tabulary}
\vspaceaftertable
\end{table}
Table~\ref{tab:uecfood256_comparison} lists the best existing results available for the UECFood256 dataset.
The depicted results show that our solution obtains the best performances by surpassing the $83\%$ and $95\%$ recognition accuracies at \textit{Top--1} and \textit{Top--5}, respectively.

More interesting are the performances obtained when no-ground truth is considered to locate the food dish within the image.
In such a case, we outperform the previous best result and obtain the overall third best \textit{Top--5} recognition accuracy, thus achieving better performance than recent methods which consider ground truth cropped images (\eg,~\cite{Hassannejad2016,He2016}).

\begin{figure*}[t]
	\centering
	\begin{subfigure}[]{\vaImVisualResWidth}
		\includegraphics[width=\linewidth]{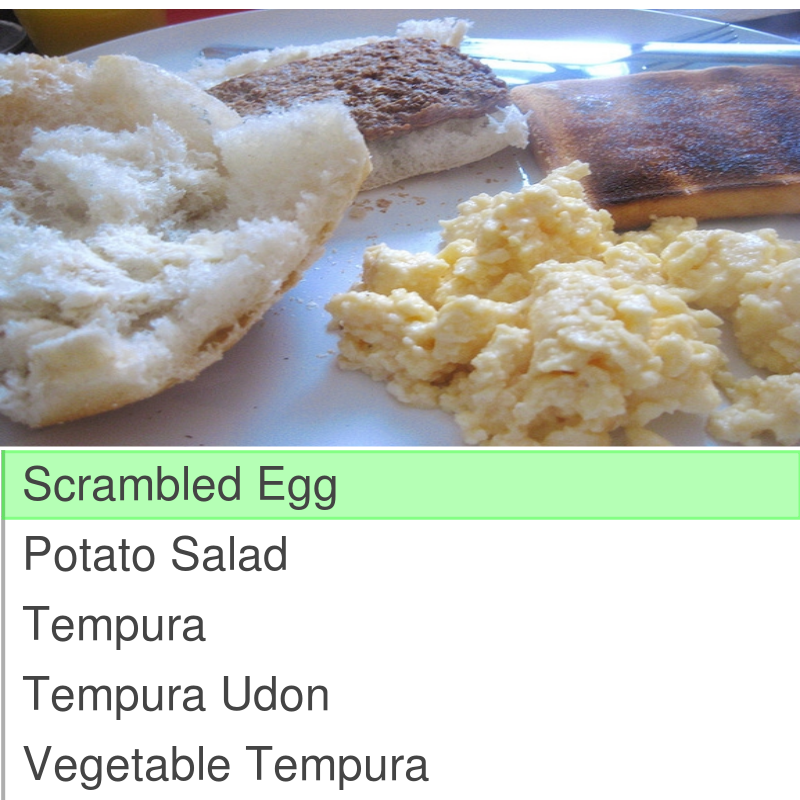}
	\end{subfigure}
	\begin{subfigure}[]{\vaImVisualResWidth}
		\includegraphics[width=\linewidth]{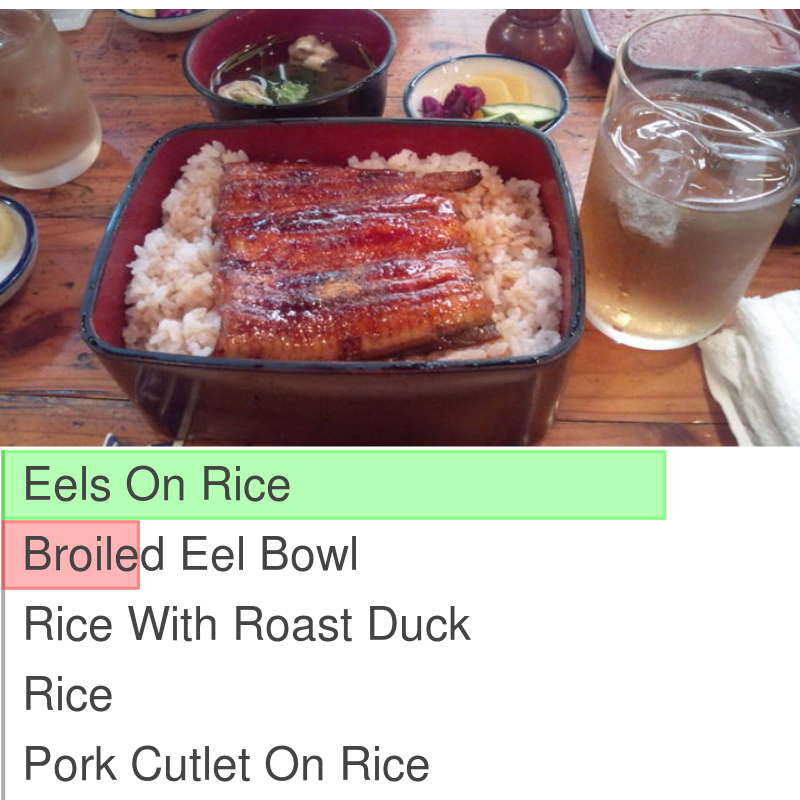}
	\end{subfigure}
	\begin{subfigure}[]{\vaImVisualResWidth}
		\includegraphics[width=\linewidth]{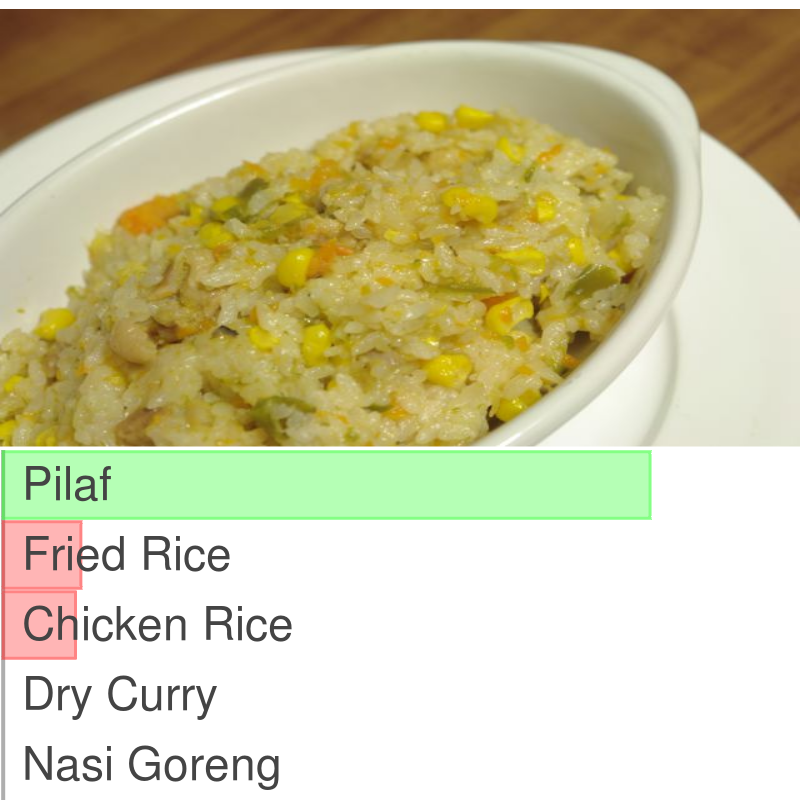}
	\end{subfigure}
	\begin{subfigure}[]{\vaImVisualResWidth}
		\includegraphics[width=\linewidth]{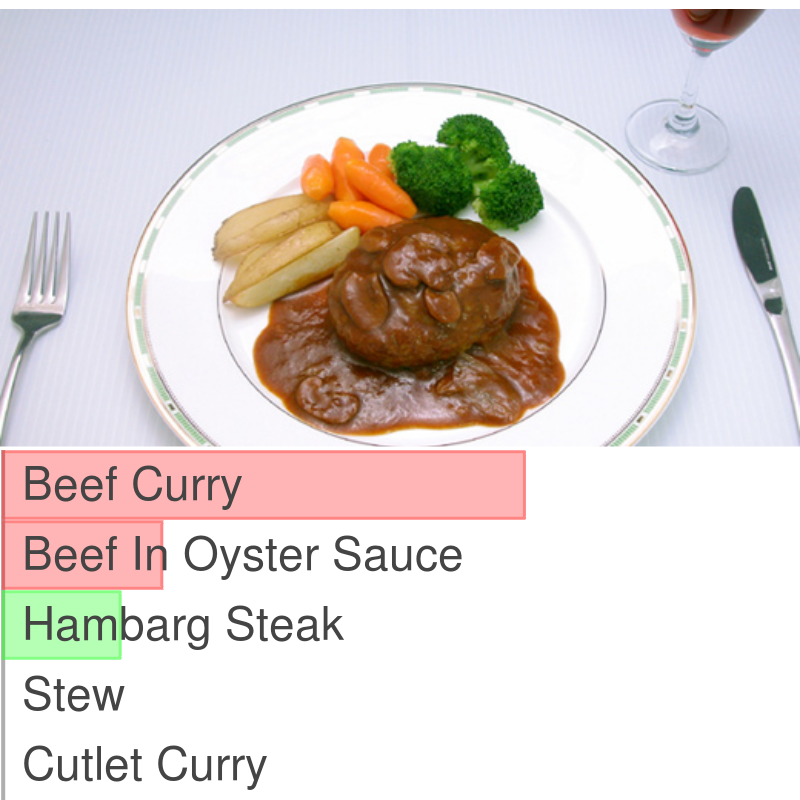}
	\end{subfigure}
	\begin{subfigure}[]{\vaImVisualResWidth}
		\includegraphics[width=\linewidth]{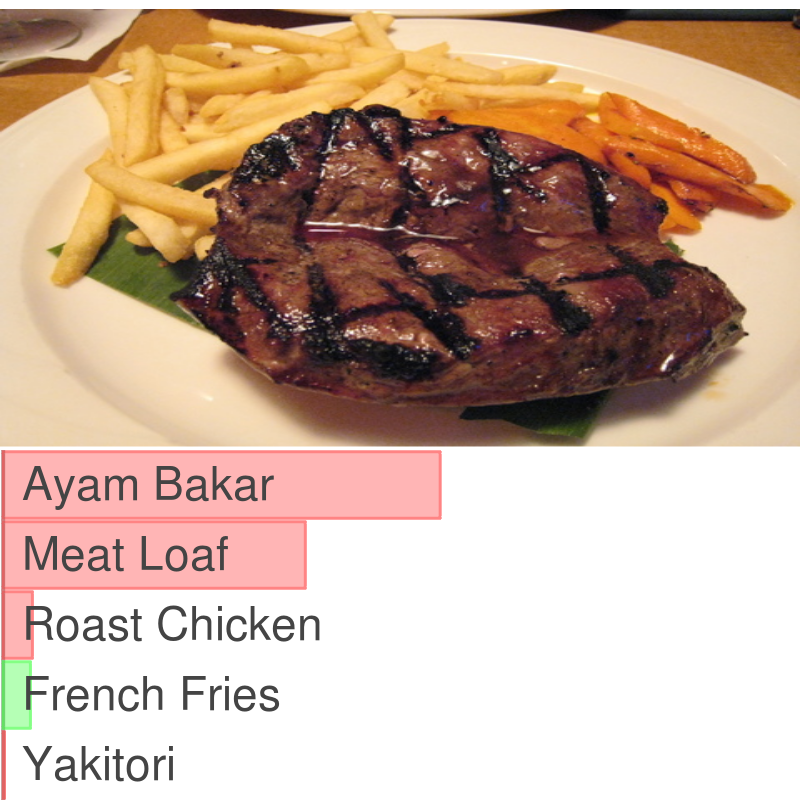}
	\end{subfigure}
	\caption{\textit{Top--5} predictions of our~\myalgoname architecture on 5 image samples from the UECFood256 dataset (with no cropped ground-truths).
		Test image are shown at the top.
		In the bar plots, predictions are ranked from top (most likely class) to bottom (less likely class).
		The true match class is represented by a green bar.
		False matches are shown with red bars.
		\textit{(Best viewed in color)}}
	\label{fig:visual_results}
	\vspaceafterfigure
\end{figure*}

Such an outcome, together with the results shown in Table~\ref{tab:uecfood100_comparison}, might indicate that our proposed solution is able to focus only on the relevant portion of an image to perform the classification task.
As shown in~\figurename~\ref{fig:visual_results}, visual inspection of the obtained performance substantiate this hypothesis.
It also demonstrate that the proposed solution has gained an high-level knowledge of the food dishes by giving high scores to food plates that are very similar to each other, or contain more than a single food class (\eg, $2^{\textrm{nd}}$ to $5^{\textrm{th}}$ sample).
To obtain more detailed insights on such results, an analysis of the visual attention performed by the architecture has been conducted (see Sec.~\ref{sec:visual_attention}).

\noindent\textbf{Food-101.}
\begin{table}[t]
\small
\centering
\caption{\textit{Top--1} and \textit{Top--5} performance on the Food-101 dataset.
	First 12 rows show the results obtained by using methods adopting hand-crafted features.
	Last 6 rows show the performance obtained by deep learning-based approaches.
Best results is highlighted in boldface font.}
\label{tab:food101_comparison}
\begin{tabulary}{1\linewidth}{L{8em} C{3em}  C{3em} | L{7em}}
\toprule
\textbf{Method} & \textbf{Top-1} & \textbf{Top-5} & \textbf{Publication} \\ \midrule
HoG & 8.85 & - & ECCV2014~\cite{Bossard2014}\\
SURF BoW-1024 & 33.47 & - & ECCV2014~\cite{Bossard2014}\\
SURF IFV-64 & 44.79 & - & ECCV2014~\cite{Bossard2014}\\
SURF IFV-64 + Color Bow-64 & 49.40& -  & ECCV2014~\cite{Bossard2014}\\
BoW & 28.51 & - & ECCV2014~\cite{Bossard2014}\\
IFV & 38.88 & - & ECCV2014~\cite{Bossard2014}\\ 
RF & 37.72 & - & ECCV2014~\cite{Bossard2014}\\
RCF & 28.46 & - & ECCV2014~\cite{Bossard2014}\\
MLDS & 42.63 & - & ECCV2014~\cite{Bossard2014}\\
RFDC & 50.76 & - & ECCV2014~\cite{Bossard2014}\\ 
SELC & 55.89 & - & CVIU2016~\cite{Martinel2016}\\ 
\hline \hline
AlexNet-CNN 		& 56.40 & - & ECCV2014~\cite{Bossard2014}\\
DCNN-FOOD 			& 70.41 & -  & ICME2015~\cite{Yanai2015a}\\
DeepFood 			& 77.4 & 93.7  & COST2016~\cite{Liu2016}\\
Inception V3 		& 88.28 & 96.88 & ECCVW2016~\cite{Hassannejad2016}\\
ResNet-200			& 88.38 & 97.85 & CVPR2016~\cite{He2016}\\
WRN					& 88.72 & 97.92 & BMVC2016~\cite{Zagoruyko2016}\\
\myalgoname 		& \textbf{90.27} & \textbf{98.71} & Proposed \\ \bottomrule
\end{tabulary}
\vspaceaftertable
\end{table}
A comparison with existing methods on the Food-101 dataset is shown in Table~\ref{tab:food101_comparison}.
Results demonstrates that our solution outperforms the best results obtained by considering hand-crafted features~\cite{Martinel2016}.
The proposed architecture performs better than all the existing deep learning-based ones by achieving a \textit{Top-1} accuracy of more than $90\%$.
Such a result show that our solution is able to learn good representations even from weakly labeled data.

\subsubsection{Ablation Analysis}
\begin{table}[t]
\small
\centering
\caption{\textit{Top--1} and \textit{Top--5} performance achieved by separately exploiting the two proposed network branches on the UECFood100, UECFood256 and Food-101 datasets.
	Slice@\myalgoname shows the results obtained using only the slice convolution branch.
	Residual@\myalgoname shows the performance achieved via the residual learning branch.}
\label{tab:ablation}
\begin{tabulary}{1\linewidth}{L{7em} | L{8em} C{3em} C{3em}}
\toprule
\textbf{Dataset} & \textbf{Model} & \textbf{Top-1} & \textbf{Top-5} \\
\midrule
\multirow{2}{*}{UECFood100} & slice@\myalgoname 	& 41.72 & 66.15	\\
							& residual@\myalgoname	& 86.71 & 98.92	\\
\midrule
\multirow{2}{*}{UECFood256} & slice@\myalgoname 	& 30.56 & 57.65	\\
							& residual@\myalgoname	& 79.76 & 93.90	\\
\midrule
\multirow{2}{*}{Food-101} 	& slice@\myalgoname  	& 46.17 & 63.57	\\
							& residual@\myalgoname 	& 88.72 & 97.92	\\							
\bottomrule
\end{tabulary}
\vspaceaftertable
\end{table}
To better understand the source of our performance, Table~\ref{tab:ablation} shows results for ablation experiments analyzing the contributions of the two architecture branches.
Specifically:
\begin{enumerate*}[label=(\roman{*})]
\item \textit{slice@\myalgoname} removes the residual network branch.
Classification is obtained by considering only the features extracted from the slice convolution branch;
\item \textit{residual@\myalgoname} performs the opposite.
Food recognition is achieved through classification of residual features.
\end{enumerate*}

Results show that for all the three datasets, the residual learning branch (\ie, \textit{resisual@\myalgoname}) largely outperforms the slice one.
We hypothesize that the reason behind such a result is due to the following facts:
\begin{enumerate*}[label=(\roman{*})]
\item The residual learning branch has been pre-trained considering a very large set of natural images (\ie, ImageNet), while the weights of the slice branch are learned from scratch.
Features extracted from a network trained on ImageNet --without fine-tuning-- have shown to be highly discriminative \textit{per se} for many visual recognition tasks~\cite{Sermanet2014,Joulin2016} (food recognition included~\cite{Martinel2016}).
Thus, it is reasonable to believe that such a pre-training introduces significant priors for the network weights.
This is confirmed by the fact that training the whole \myalgoname architecture from scratch results in a recognition rate of $78.12\%$, $68.37\%$, and $79.45\%$ on the three considered datasets, respectively.
\item Capturing only vertical layer features excludes learning of other distinctive traits that are not specific of vertically structured food dishes.
\end{enumerate*}

\subsection{Visual Attention}
\label{sec:visual_attention}
\begin{figure}[t]
	\centering
	\begin{subfigure}[]{\vaImWidth}
		\includegraphics[width=\linewidth, height=7.6em]{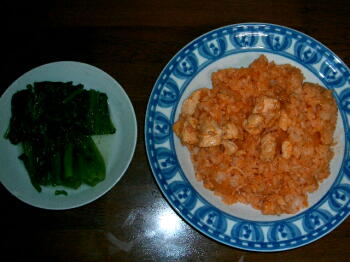} 
	\end{subfigure}
	\begin{subfigure}[]{\vaImWidth}
		\includegraphics[width=\linewidth]{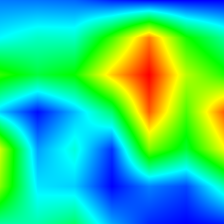}
	\end{subfigure}
	\begin{subfigure}[]{\vaImWidth}
		\includegraphics[width=\linewidth]{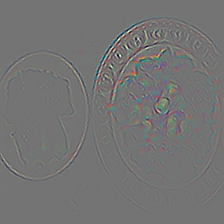}
	\end{subfigure}
	\\ \vspace{0.1em}
	\begin{subfigure}[]{\vaImWidth}
		\includegraphics[width=\linewidth, height=7.6em]{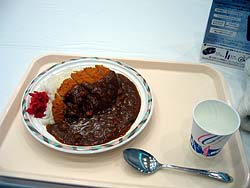} 
	\end{subfigure}
	\begin{subfigure}[]{\vaImWidth}
		\includegraphics[width=\linewidth]{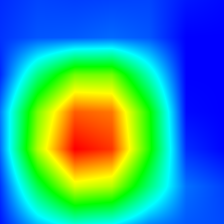}
	\end{subfigure}
	\begin{subfigure}[]{\vaImWidth}
		\includegraphics[width=\linewidth]{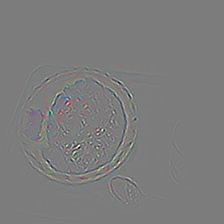}
	\end{subfigure}
	\caption{Analysis of the visual attention obtained through our architecture on two randomly selected images from the UECFood100 dataset.
		First column is the input image, second column shows the visual attention with a color-coded plot (blue means lower attention, red higher).
		Last column depicts the gradient computed with respect to the input image, showing that features are extracted only from the relevant image region.
		Results obtained through Guided Grad-CAM visual explanation~\cite{Selvaraju2016}.		
		\textit{(Best viewed in color)}}
	\label{fig:visual_attention}
	\vspaceafterfigure
\end{figure}
Results on the UECFood100 and UECFoo256 datasets show that our approach can extract the useful information for classification from the relevant image regions only.
To have a richer grasp on this outcome, we have conducted an analysis of the visual attention performed by the architecture.
Towards this end, we have exploited the recent Grad-CAM approach~\cite{Selvaraju2016}.
It allows to obtain a coarse localization map regarding the important regions in the image which are considered for classification.

As show in~\ref{fig:visual_attention}, when more than a single food dish is present in the image, the~\myalgoname architecture is able to focus only on the image portion that contains the object of interest.
This is substantiated by the fact that features are not extracted from other non-relevant food/non-food objects (\eg, the plate containing green leaves, the spoon, the paper glass, \etc).

\subsection{Discussion}
\noindent\textbf{Outcomes.}
Results obtained for the three datasets demonstrate that:
\begin{enumerate*}[label=(\roman{*})]
\item our solution shows better performance with respect to existing approaches either based on hand-crafted features or on deep learning schemes;
\item while the residual learning branch brings most of the classification power, by combining the so extracted features with the vertical layer traits discovered through the slice convolution branch the best achievements are attained;
\item the proposed~\myalgoname architecture is able to self-discover the image portion that should be considered to extract the features, hence to emit the classification.
\end{enumerate*}

This shows that our approach is able to well address the many non-trivial challenges in food recognition and is not designed to tackle the specific problems brought in by a single dataset.

\noindent\textbf{Limitations.}
It is a matter of fact that nowadays food recognition algorithms are very attractive for mobile platforms.
Our solution require substantial memory loads as well as significant computational efforts to process a single datum, thus denying a possible deployment of our approach on these devices.
A possible solution to such a problem would be to compress the network to obtain a shallower architecture~\cite{Han2016} or to exploit binary weights~\cite{Rastegari2016}.
We demand this study to future works.

\section{Conclusion}
In this paper, a system for automatic food recognition based on a deep learning solution specifically designed for the considered task has been proposed.
The~\myalgoname architecture combines features extracted from two main network branches.
The residual learning branch provides a deep hierarchy which is able to capture the food traits of the majority of the existing food categories.
The slice convolution branch captures the vertical layers of the food dishes which present such a peculiarity.
The features extracted from these branches are fused then exploited to emit the classification.

To demonstrate the benefits of the proposed solution, evaluations on three benchmark datasets have been conducted.
Comparisons with existing methods have shown that by exploiting both the architecture branches together better performance than state-of-the-art approaches are achieved regardless the considered dataset.
The visual attention analysis has shown that the network is able to self-identify the relevant portions of the image that should be considered for classification.

{\small
\bibliography{egbib}

\begin{thebibliography}{10}\itemsep=-1pt

\bibitem{Bettadapura2015a}
V.~Bettadapura, E.~Thomaz, A.~Parnami, G.~D. Abowd, and I.~Essa.
\newblock {Leveraging Context to Support Automated Food Recognition in
  Restaurants}.
\newblock In {\em Winter Conference on Applications of Computer Vision}, 2015.

\bibitem{Bossard2014}
L.~Bossard, M.~Guillaumin, and L.~{Van Gool}.
\newblock {Food-101 – Mining Discriminative Components with Random Forests}.
\newblock In {\em European Conference Computer Vision}, 2014.

\bibitem{Chen2016}
J.~Chen and C.-W. Ngo.
\newblock {Deep-based Ingredient Recognition for Cooking Recipe Retrieval}.
\newblock In {\em ACM Multimedia}, pages 1----6, 2016.

\bibitem{Deng2016}
Z.~Deng, A.~Vahdat, H.~Hu, and G.~Mori.
\newblock {Structure Inference Machines: Recurrent Neural Networks for
  Analyzing Relations in Group Activity Recognition}.
\newblock In {\em International Conference on Computer Vision and Pattern
  Recognition}, 2016.

\bibitem{Farinella2014}
G.~M. Farinella, D.~Allegra, and F.~Stanco.
\newblock {A Benchmark Dataset to Study the Representation of Food Images}.
\newblock In {\em European Conference Computer Vision Workshops}, 2014.

\bibitem{Farinella2014a}
G.~M. Farinella, M.~Moltisanti, and S.~Battiato.
\newblock {Classifying Food Images Represented as Bag of Textons}.
\newblock In {\em International Conference on Image Processing}, pages
  5212----5216, 2014.

\bibitem{Feichtenhofer2016}
C.~Feichtenhofer, A.~Pinz, and A.~Zisserman.
\newblock {Convolutional Two-Stream Network Fusion for Video Action
  Recognition}.
\newblock In {\em International Conference on Computer Vision and Pattern
  Recognition}, pages 1933--1941, 2016.

\bibitem{Han2016}
S.~Han, H.~Mao, and W.~J. Dally.
\newblock {Deep compression: Compressing deep neural network with pruning,
  trained quantization and huffman coding}.
\newblock In {\em International Conference on Learning Representations}, pages
  1--13, 2016.

\bibitem{Hassannejad2016}
H.~Hassannejad, G.~Matrella, P.~Ciampolini, I.~{De Munari}, M.~Mordonini, and
  S.~Cagnoni.
\newblock {Food Image Recognition Using Very Deep Convolutional Networks}.
\newblock In {\em European Conference Computer Vision Workshops and
  Demonstrations}, pages 41--49, 2016.

\bibitem{He2015}
K.~He, X.~Zhang, S.~Ren, and J.~Sun.
\newblock {Deep Residual Learning for Image Recognition}.
\newblock {\em ArXiv e-prints}, 1512.03385, 2015.

\bibitem{He2016}
K.~He, X.~Zhang, S.~Ren, and J.~Sun.
\newblock {Deep Residual Learning for Image Recognition}.
\newblock In {\em International Conference on Computer Vision and Pattern
  Recognition}, pages 770----778, 2016.

\bibitem{He2016a}
K.~He, X.~Zhang, S.~Ren, and J.~Sun.
\newblock {Identity Mappings in Deep Residual Networks}.
\newblock In {\em European Conference on Computer Vision}, pages 630--645,
  2016.

\bibitem{Hendricks2016}
L.~A. Hendricks, S.~Venugopalan, M.~Rohrbach, R.~Mooney, K.~Saenko, and
  T.~Darrell.
\newblock {Deep Compositional Captioning: Describing Novel Object Categories
  without Paired Training Data}.
\newblock In {\em International Conference on Computer Vision and Pattern
  Recognition}, pages 1--10, 2016.

\bibitem{Howard2013}
A.~G. Howard.
\newblock {Some Improvements on Deep Convolutional Neural Network Based Image
  Classification}.
\newblock {\em ArXiv e-prints}, 1312.5402, 2013.

\bibitem{Hu2015}
R.~Hu, H.~Xu, M.~Rohrbach, J.~Feng, K.~Saenko, and T.~Darrell.
\newblock {Natural Language Object Retrieval}.
\newblock In {\em International Conference on Computer Vision and Pattern
  Recognition}, pages 4555--4564, 2016.

\bibitem{Joulin2016}
A.~Joulin, L.~van~der Maaten, A.~Jabri, and N.~Vasilache.
\newblock {Learning Visual Features from Large Weakly Supervised Data}.
\newblock In {\em European Conference on Computer Vision}, pages 67--84, 2016.

\bibitem{Kawano2013}
Y.~Kawano and K.~Yanai.
\newblock {Real-Time Mobile Food Recognition System}.
\newblock {\em Computer Vision and Pattern Recognition Workshops}, pages 1--7,
  2013.

\bibitem{Kawano2014c}
Y.~Kawano and K.~Yanai.
\newblock {Automatic expansion of a food image dataset leveraging existing
  categories with domain adaptation}.
\newblock In {\em European Conference Computer Vision Workshops and
  Demonstrations}, pages 3--17, 2014.

\bibitem{Kawano2014b}
Y.~Kawano and K.~Yanai.
\newblock {Food Image Recognition with Deep Convolutional Features}.
\newblock In {\em ACM International Joint Conference on Pervasive and
  Ubiquitous Computing (UbiComp)}, pages 589--593, 2014.

\bibitem{Kawano2014d}
Y.~Kawano and K.~Yanai.
\newblock {FoodCam-256: A Large-scale Real-time Mobile Food RecognitionSystem
  employing High-Dimensional Features and Compression of Classifier Weights}.
\newblock In {\em ACM International Conference on Multimedia}, pages 761--762,
  2014.

\bibitem{Kawano2014}
Y.~Kawano and K.~Yanai.
\newblock {FoodCam: A real-time mobile food recognition system employing Fisher
  Vector}.
\newblock {\em Multimedia Tools and Applications}, pages 369--373, 2014.

\bibitem{Kong2012}
F.~Kong and J.~Tan.
\newblock {DietCam: Automatic dietary assessment with mobile camera phones}.
\newblock {\em Pervasive and Mobile Computing}, 8(1):147--163, 2012.

\bibitem{Krizhevsky2012a}
A.~Krizhevsky, I.~Sutskever, and G.~Hinton.
\newblock {ImageNet Classification with Deep Convolutional Neural Networks}.
\newblock In {\em Advances in Neural Information Processing Systems}, pages
  1097----1105, 2012.

\bibitem{Liu2016}
C.~Liu, Y.~Cao, Y.~Luo, G.~Chen, V.~Vokkarane, and Y.~Ma.
\newblock {Deepfood: Deep learning-based food image recognition for
  computer-aided dietary assessment}.
\newblock In {\em IEEE International Conference on Smart Homes and Health
  Telematics}, volume 9677, pages 37--48, 2016.

\bibitem{Martinel2016}
N.~Martinel, C.~Piciarelli, and C.~Micheloni.
\newblock {A supervised extreme learning committee for food recognition}.
\newblock {\em Computer Vision and Image Understanding}, 148:67--86, jul 2016.

\bibitem{Matsuda2012}
Y.~Matsuda, H.~Hoashi, and K.~Yanai.
\newblock {Recognition of multiple-food images by detecting candidate regions}.
\newblock In {\em International Conference on Multimedia and Expo}, pages
  25--30, 2012.

\bibitem{Myers2016}
A.~Myers, N.~Johnston, V.~Rathod, A.~Korattikara, A.~Gorban, N.~Silberman,
  S.~Guadarrama, G.~Papandreou, J.~Huang, and K.~Murphy.
\newblock {Im2Calories : towards an automated mobile vision food diary}.
\newblock In {\em International Conference on Computer Vision}, 2016.

\bibitem{who2015fs}
W.~H. Organization.
\newblock {Obesity and overweight - fact sheet n. 311}.
\newblock 2015.

\bibitem{Qi2014}
X.~Qi, R.~Xiao, C.-G. Li, Y.~Qiao, J.~Guo, and X.~Tang.
\newblock {Pairwise Rotation Invariant Co-occurrence Local Binary Pattern}.
\newblock {\em IEEE Transactions on Pattern Analysis and Machine
  Iintelligence}, 36(11):2199 -- 2213, 2014.

\bibitem{Rastegari2016}
M.~Rastegari, V.~Ordonez, J.~Redmon, and A.~Farhadi.
\newblock {XNOR-Net: ImageNet Classification Using Binary Convolutional Neural
  Networks}.
\newblock In {\em European Conference on Computer Vision}, pages 525--542,
  2016.

\bibitem{Russakovsky2015}
O.~Russakovsky, J.~Deng, H.~Su, J.~Krause, S.~Satheesh, S.~Ma, Z.~Huang,
  A.~Karpathy, A.~Khosla, M.~Bernstein, A.~C. Berg, and L.~Fei-Fei.
\newblock {ImageNet Large Scale Visual Recognition Challenge}.
\newblock {\em International Journal of Computer Vision}, 115(3):211--252, dec
  2015.

\bibitem{Sanchez2013}
J.~Sanchez, F.~Perronnin, T.~Mensink, and J.~Verbeek.
\newblock {Image Classification with the Fisher Vector: Theory and Practice}.
\newblock {\em International Journal of Computer Vision}, 105(3):222--245,
  2013.

\bibitem{Selvaraju2016}
R.~R. Selvaraju, A.~Das, R.~Vedantam, M.~Cogswell, D.~Parikh, and D.~Batra.
\newblock {Grad-CAM: Why did you say that? Visual Explanations from Deep
  Networks via Gradient-based Localization}.
\newblock {\em ArXiv e-prints}, 1610.02391, 2016.

\bibitem{Sermanet2014}
P.~Sermanet, D.~Eigen, X.~Zhang, M.~Mathieu, R.~Fergus, and Y.~LeCun.
\newblock {OverFeat: Integrated Recognition , Localization and Detection using
  Convolutional Networks}.
\newblock In {\em International Conference on Learning Representations}, pages
  1--15, 2014.

\bibitem{Simonyan2015}
K.~Simonyan and A.~Zisserman.
\newblock {Very Deep Convolutional Networks for Large-Scale Image Recognition}.
\newblock {\em International Conference on Learning Representations}, pages
  1--14, 2015.

\bibitem{Srivastava2015}
R.~K. Srivastava, K.~Greff, and J.~Schmidhuber.
\newblock {Highway Networks}.
\newblock {\em ArXiv e-prints}, 1505.00387, 2015.

\bibitem{Szegedy2015}
C.~Szegedy, {Wei Liu}, {Yangqing Jia}, P.~Sermanet, S.~Reed, D.~Anguelov,
  D.~Erhan, V.~Vanhoucke, and A.~Rabinovich.
\newblock {Going deeper with convolutions}.
\newblock In {\em International Conference on Computer Vision and Pattern
  Recognition}, pages 1--9. IEEE, jun 2015.

\bibitem{Yanai2015a}
K.~Yanai and Y.~Kawano.
\newblock {Food image recognition using deep convolutional network with
  pre-training and fine-tuning}.
\newblock In {\em International Conference on Multimedia {\&} Expo Workshops},
  pages 1--6. IEEE, jun 2015.

\bibitem{Yang2010}
S.~Yang, M.~Chen, D.~Pomerleau, and R.~Sukthankar.
\newblock {Food recognition using statistics of pairwise local features}.
\newblock {\em International Conference on Computer Vision and Pattern
  Recognition}, pages 2249--2256, 2010.

\bibitem{Zagoruyko2016}
S.~Zagoruyko and N.~Komodakis.
\newblock {Wide Residual Networks}.
\newblock In {\em British Machine Vision Conference}, 2016.

\bibitem{Zhang2015a}
Z.~Zhang, S.~Fidler, and R.~Urtasun.
\newblock {Instance-Level Segmentation with Deep Densely Connected MRFs}.
\newblock In {\em International Conference on Computer Vision and Pattern
  Recognition}, pages 669--677, 2015.

\bibitem{Zhu2010}
F.~Zhu, M.~Bosch, I.~Woo, S.~Kim, C.~J. Boushey, D.~S. Ebert, and E.~J. Delp.
\newblock {The Use of Mobile Devices in Aiding Dietary Assessment and
  Evaluation.}
\newblock {\em IEEE journal of selected topics in signal processing},
  4(4):756--766, 2010.

\bibitem{Zhu2016}
Y.~Zhu, O.~Groth, M.~Bernstein, and L.~Fei-Fei.
\newblock {Visual7W: Grounded Question Answering in Images}.
\newblock In {\em International Conference on Computer Vision and Pattern
  Recognition}, pages 4995--5004, 2016.

\end{thebibliography}
\bibliographystyle{ieee}
}

\end{document}